\pdfoutput=1

\documentclass[11pt]{article}

\usepackage[final]{acl}

\usepackage{times}
\usepackage{latexsym}

\usepackage[T1]{fontenc}

\usepackage[utf8]{inputenc}

\usepackage{microtype}

\usepackage{inconsolata}

\usepackage{graphicx}

\usepackage{times}
\usepackage{soul}
\usepackage{url}
\usepackage{amsmath,amssymb,amsfonts}
\usepackage{booktabs}
\usepackage{algorithm}
\usepackage{algorithmic}
\usepackage[switch]{lineno}

\usepackage{subfigure}
\usepackage{color}
\usepackage{xcolor}
\usepackage{multirow}
\usepackage{makecell}

\title{

FedCoT: Federated Chain-of-Thought Distillation for Large Language Models 

}

\author{
 \textbf{Tao Fan\textsuperscript{1, 2}},
 \textbf{Weijing Chen\textsuperscript{2}},
 \textbf{Yan Kang\textsuperscript{2}},
 \textbf{Guoqiang Ma\textsuperscript{2}},
 \\
 \textbf{Hanlin Gu\textsuperscript{2}},
 \textbf{Yuanfeng Song\textsuperscript{2}},
 \textbf{Lixin Fan\textsuperscript{2}},
 \textbf{Qiang Yang\textsuperscript{3}}
\\
 \textsuperscript{1} Hong Kong University of Science and Technology, Hong Kong, China
 \\
 \textsuperscript{2} WeBank, China
\\
 \textsuperscript{3}Hong Kong Polytechnic University, Hong Kong, China
\\
 \small{
   \textbf{Correspondence:} \href{mailto:tfanac@cse.ust.hk}{tfanac@cse.ust.hk}, \href{mailto:qyang@cse.ust.hk}{qyang@cse.ust.hk}
 }
}

\begin{document}
\maketitle

\begin{abstract}
Large Language Models (LLMs) have emerged as a transformative force in artificial intelligence, demonstrating exceptional proficiency across various tasks. However, their deployment in resource-constrained environments and concerns over user data privacy pose significant challenges. In contrast, Small Language Models (SLMs) offer computational efficiency but often lag in performance. To address these issues, we propose FedCoT, a federated framework designed for the Chain-of-Thought (CoT) distillation of knowledge from LLMs to SLMs, while ensuring the preservation of clients' data privacy. FedCoT ensures secure and efficient knowledge transfer from an LLM on a high-powered server to an SLM on a resource-constrained client, while adhering to privacy requirements. Leveraging perturbed prompts and rationales generated through the CoT approach, the framework enhances the performance of the client's SLM without compromising user data privacy within a multi-task learning framework. We propose two privacy protection strategies: the Exponential Mechanism Strategy and the Adaptive Exponential Mechanism Strategy, which balance user prompt privacy and the usability of rationales. Empirical evaluation on various text generation tasks demonstrates the effectiveness of FedCoT in training task-specific SLMs with enhanced performance while prioritizing data privacy protection. Our code has been contributed to the FATE open-source project and is now publicly accessible at \textit{\url{https://github.com/FederatedAI/FATE-LLM/tree/main/python/fate_llm/algo/fedcot}}

\end{abstract}

\section{Introduction} 

Large Language Models (LLMs) have risen as a revolutionary force in artificial intelligence. Prominent LLMs, such as GPT-4~\cite{Gpt-4}, LLaMA~\cite{touvron2023llama}, and Qwen~\cite{bai2023qwen},  have garnered the attention of researchers and practitioners alike, demonstrating unparalleled proficiency across numerous tasks. Nevertheless, the sheer size of these models presents significant obstacles for real-world deployment, particularly in environments with limited resources~\cite{fan2025ten, fan2025fedmkt, fan2023fate-llm, kang2023grounding}. Meanwhile, as LLMs gain escalating popularity and widespread utilization, privacy concerns have moved to the forefront, especially when it comes to user data and LLMs inference. In contrast, Small Language Models (SLMs) often exhibit superior computational efficiency and faster convergence rates, rendering them perfectly suited for real-time applications or resource-constrained environments. Nonetheless, SLMs also possess certain drawbacks stemming from their performance limitations. The question then arises: \textit{How can we effectively combine the predictive prowess of LLMs with the nimbleness of SLMs, all while adhering to privacy requirements?} 

To address these challenges, we propose FedCoT, a federated framework designed for the \textit{Chain-of-Thought (CoT)}~\cite{wei2022chain} distillation of knowledge from LLMs to SLMs, while ensuring the preservation of clients' data privacy. 
FedCoT ensures secure and efficient knowledge transfer from an LLM on a high-powered server to an SLM on a resource-constrained client.
The challenge lies in maintaining the privacy of client data while leveraging the server's LLM to aid in training the client's SLM for text generation tasks, thereby elevating its performance. FedCoT aims to bridge this gap, enabling secure and efficient knowledge transfer between LLM and SLM, and ultimately enhancing the capabilities of the SLM without compromising privacy.

\begin{figure*}[ht]
 \centering
\subfigure[Overview of our proposed \textbf{FedCoT} framework.]{
        \includegraphics[width=0.55\textwidth]{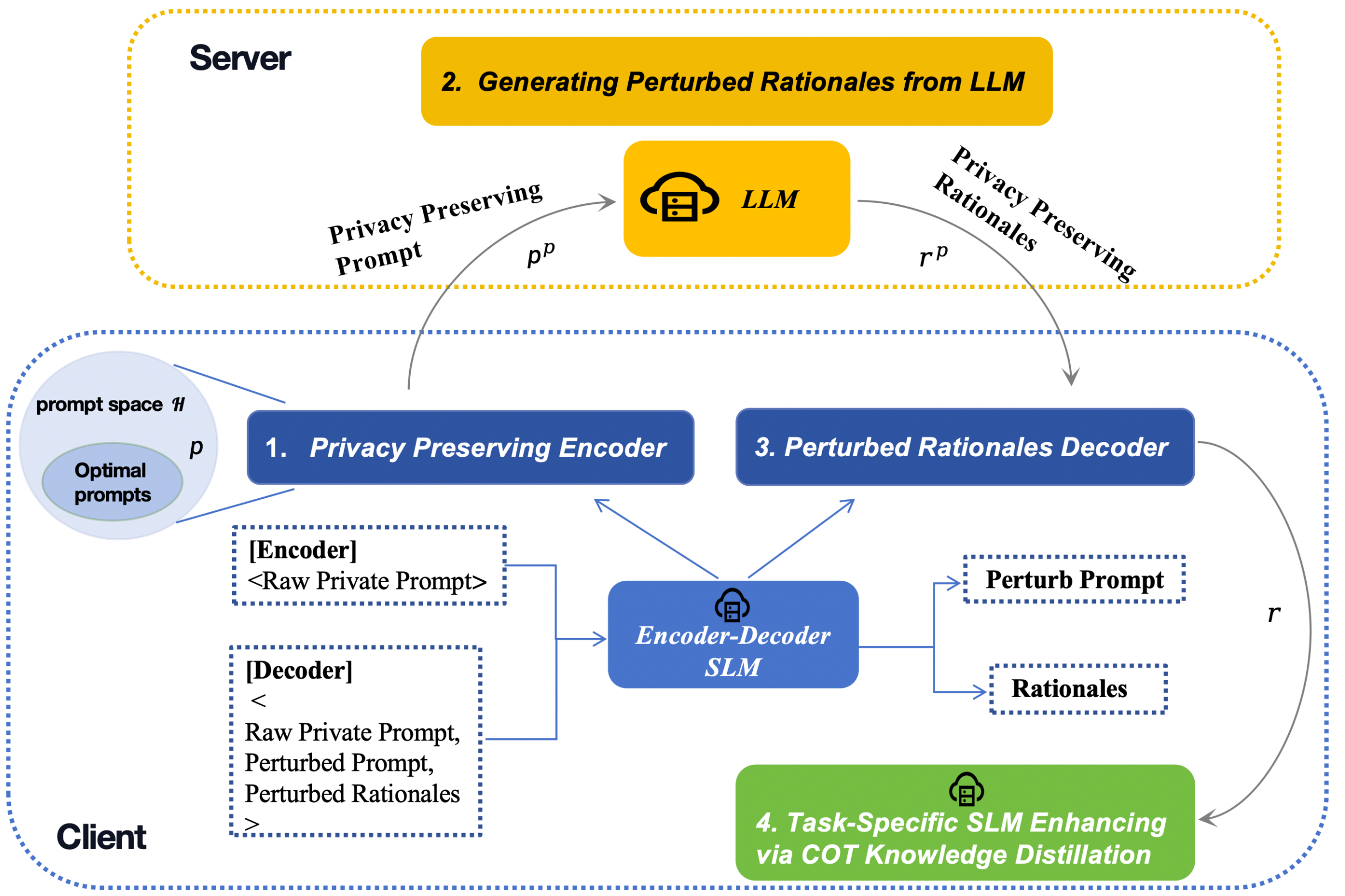} 
        \label{fig:fedcot}
    }
     \hfill
    \subfigure[Privacy-Preserving Rationals Generation.]{
        \includegraphics[width=0.4\textwidth]{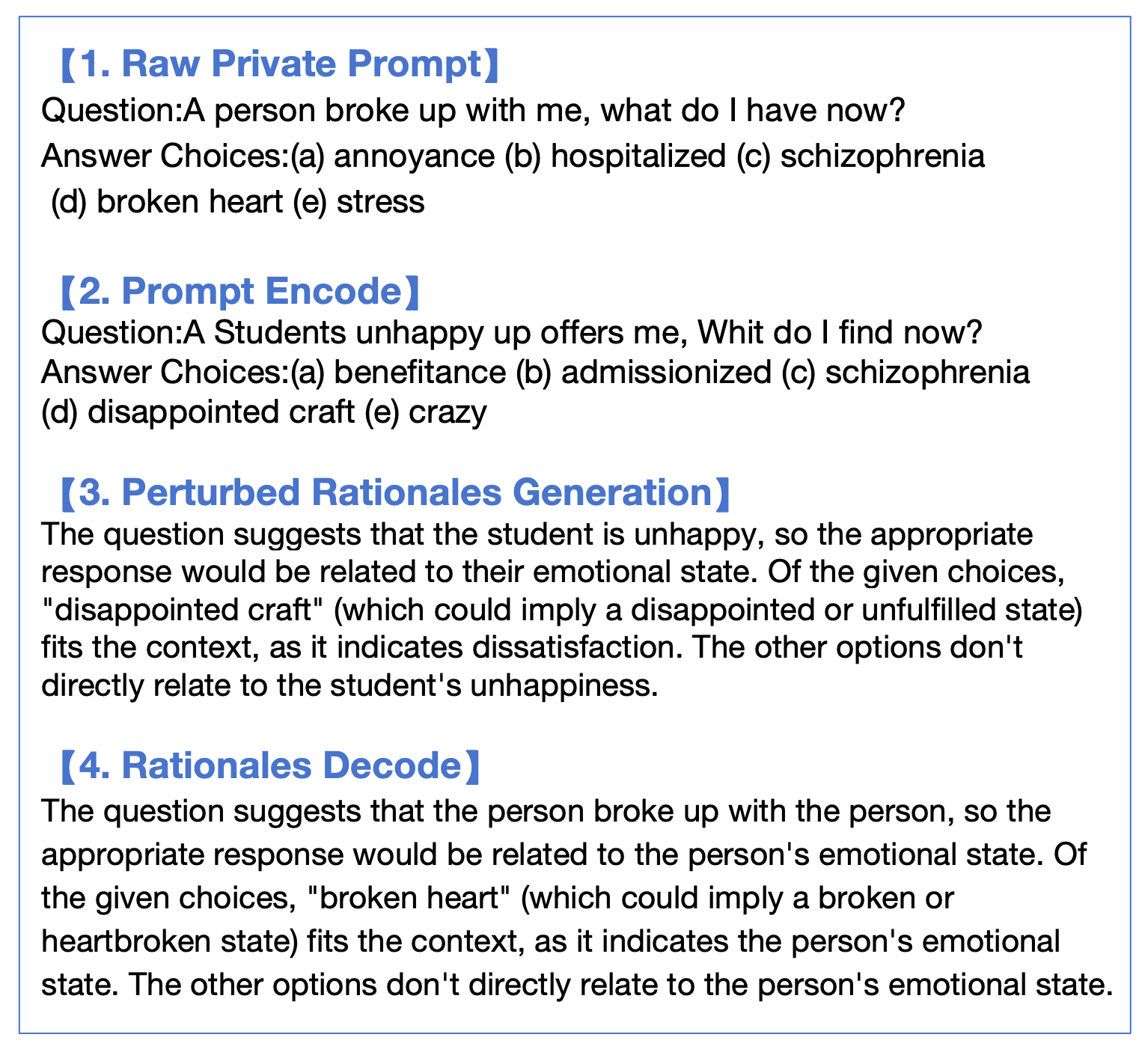} 
       \label{fig:example}
    }

    \caption{The overview of our proposed \textbf{FedCoT}. The FedCoT comprises four key components: (1) The \textit{Prompt Encoder}, which perturbs user prompts to ensure privacy;  (2) The \textit{LLM}, generating perturbed rationales based on the perturbed prompts; (3) The \textit{Perturbed Rationales Decoder}, which decodes the perturbed rationales back into a usable form;  (4) The \textit{Task-Specific SLM Enhancing via CoT Knowledge Distillation}, utilizing both original labeled data and filtered rationales data for  multi-task learning.}
    \label{fig:fedcot&example}

\end{figure*}

As illustrated in Figure \ref{fig:fedcot}, within our framework, the process works as follows. 
Initially, the client transmits perturbed prompts to the server's LLM. 
These prompts are protected by the FedCoT prompt encoder, which employs Differential Privacy (DP) principles~\cite{dwork2006differential,mcsherry2007mechanism}, ensuring privacy protection.
Subsequently, the server's LLM generates perturbed rationales from these prompts through the CoT approach and relays them back to the client. Upon receiving these perturbed rationales, the client's rationales decoder reconstructs them into their original, aligned form corresponding to the raw prompt.
Ultimately, the client utilizes CoT knowledge distillation~\cite{hsieh2023distilling,li2023symbolic}
to train its \textit{Task-Specific SLM}. This process leverages both label data and rationales within a multi-task learning paradigm~\cite{wei2022chain,hsieh2023distilling,zhang2021survey}.
These rationales justify the predicted labels and serve as insightful guidance for training smaller and domain-specific models.

Previous endeavors to incorporate DP into language models, specifically through DP-SGD~\cite{song2013stochastic}, have primarily centered on navigating the delicate balance between utility and privacy. This is achieved by introducing calibrated noise into gradients or text representations during the model training process. Nonetheless, these methods inherently rely on a trusted server to gather data from data owners for model training~\cite{chen-etal-2023-customized}, significantly limiting their applicability in scenarios where such trusted servers are not available, as is the case in our research context.

Within the FedCoT framework, to achieve a balance between preserving the privacy of user prompts and enhancing the usability of rationales, 
we introduce two privacy protection strategies: the \textit{Exponential Mechanism  Strategy} and the \textit{Adaptive Exponential Mechanism Strategy}. 
In the \textit{Exponential Mechanism  Strategy}, we utilize an exponential mechanism to obfuscate the  prompts~\cite{mcsherry2007mechanism,yue2021differential,chen-etal-2023-customized},  followed by decoding the perturbed rationales through In-Context Learning (ICL)~\cite{dong2024survey,tong2025inferdpt}.
In the \textit{Adaptive Exponential Mechanism Strategy}, we utilize an  Encoder-Decoder SLM specifically designed to encode original prompts into perturbed prompts and subsequently decode perturbed rationales back into their original form. To effectively train this unified Encoder-Decoder SLM, we utilize a multi-task learning paradigm~\cite{zhang2021survey}, encompassing both the encoding and decoding training processes.

Our contributions are summarized as follows:
\begin{itemize}
\item 
\textbf{Federated Framework for CoT Distillation in LLMs}. We propose FedCoT, a novel federated framework that facilitates secure and efficient knowledge transfer from LLM to SLM in resource-constrained environments. FedCoT leverages CoT knowledge distillation to enhance Task-Specific SLM within the client. This process leverages rationales produced by the LLM on the server, thereby enriching the client-side SLMs with valuable task-related knowledge.

\item  

\textbf{Privacy as a Priority.} FedCoT leverages an \textit{Adaptive Exponential Mechanism Strategy} tailored for encoding prompt to ensure their obfuscation and decoding perturbed rationales. The strategies effectively balance user prompt privacy and the usability of rationales.

\item 
\textbf{Empirical Evaluation and Enhanced Performance of Task-Specific SLM}. Through experiments on various text generation tasks, FedCoT demonstrates the effectiveness of its framework in training task-specific SLM with enhanced performance. By harnessing the rationales generated by the server-side LLM, FedCoT provides valuable task-specific knowledge to the SLM.

\end{itemize}

\section{Related Work}

\subsection{Differential Privacy}
\label{sec:dp}
In this section, We briefly revisit two important definitions of differential privacy: $\epsilon$-Differential Privacy and Exponential Mechanism (EM).

\textbf{$\epsilon$-Differential Privacy.} Differential privacy (DP) ~\cite{dwork2006differential} is a rigorous mathematical framework that provides strong privacy guarantees for data analysis. It ensures that the output of an algorithm remains statistically indistinguishable whether a particular individual's data is included or excluded from the dataset. Formally, a randomized mechanism $ M $ provides $\epsilon$-differential privacy if for all neighboring datasets $ D $ and $ D' $ (differing in at most one record) and for all sets $ S $ of possible outputs:

\begin{equation}\label{eq:dp-1}
\begin{aligned}
\Pr[M(D) \in S] \leq e^\epsilon \Pr[M(D') \in S]
\end{aligned}
\end{equation}
where $\epsilon$ is the privacy budget that controls the level of indistinguishability.

\textbf{Exponential Mechanism.} The Exponential Mechanism ~\cite{mcsherry2007mechanism} allows for the selection of an outcome from a set of possible outcomes with probabilities proportional to the exponential of their utility scores. Formally, given a utility function $ u: D \times R \rightarrow \mathbb{R} $ that maps each dataset $ D $ and possible outcome $ r $ to a real-valued score, the Exponential Mechanism $ M(D, u, R) $ satisfies $\epsilon$-differential privacy if it selects and outputs an $ r \in R $ with probability:

\begin{equation}\label{eq:dp-2}
\begin{aligned}
\Pr[M(D) = r] \propto \exp\left(\frac{\epsilon u(D, r)}{2 \Delta u}\right)
\end{aligned}
\end{equation}
where $\Delta u$ is the sensitivity of the utility function (in our work, we use cosine similarity as the utility function), defined as the maximum change in utility score when a single record is added or removed from the dataset:

\begin{equation}\label{eq:dp-3}
\begin{aligned}
\Delta u = \max_{D, D', r} |u(D, r) - u(D', r)|
\end{aligned}
\end{equation}

\subsection{Chain of Thought in Large Language Models}
The Chain of Thought (CoT) approach has recently garnered significant attention in the realm of LLMs, thanks primarily to its remarkable ability to enhance the reasoning capabilities of these models. This innovative concept was first introduced by ~\cite{wei2022chain}. Their research demonstrated that by prompting LLMs to produce a sequence of intermediary reasoning steps (rationales), the models' performance in handling intricate reasoning tasks could be notably boosted. 
Since the introduction of CoT, several studies have delved into its extensions and variations. For example, ~\cite{kojima2022large} proposed the use of zero-shot CoT, where the model is prompted to generate rationales without relying on prior examples.
CoT has also been applied to various domains, including arithmetic reasoning~\cite{cobbe2021training}, commonsense reasoning~\cite{klein2020contrastive}. 
Recent studies by~\cite{hsieh2023distilling,ho2023large,li2023symbolic},  have capitalized on the generated rationales as a form of insightful supervision to train smaller and domain-specific models. 
\textbf{\textit{However, previous studies have not addressed the domain-specific data privacy issue that arises when LLMs and domain-specific smaller models are deployed across different parties. In our work, we endeavor to address this significant challenge.}}

\section{The Proposed FedCoT Framework}

In this section, we introduce FedCoT, 
a  federated framework designed for the \textit{CoT} distillation of knowledge from LLMs hosted on a high-powered server to SLMs deployed on a resource-constrained client. The FedCoT framework can enhance the performance of SLMs while maintaining client data's privacy, leveraging the capabilities of LLM. We assume the server to be \textit{semi-honest}, implying that it may attempt to recover the private data of the client from the information it observes.
We illustrate the FedCoT in Figure~\ref{fig:fedcot}, outline its training algorithm in Algorithm~\ref{alg:fedcot}, and detail its resource requirements in Appendix~\ref{sec:appendix-requirement}.

\begin{algorithm}[h] 
\caption{FedCoT}
\label{alg:fedcot}

\textbf{Input:} \\
    $T$: total number of rounds; \\
    $\mathcal{P}$: encoding training datasets; \\
    $\mathcal{R}$: decoding training datasets; \\
    $\mathcal{D}$: task-specific training datasets; \\
    $\eta_\phi$: learning rate of Encoder-Decoder SLM; \\
    $\eta_\omega$: learning rate of Task-Specific SLM.
  
\textbf{Output:} $g_{\phi}$, $f_{\omega}$.

\begin{algorithmic}[1]
  \STATE \textcolor{gray}{$\triangleright$ Multi-Task Training for Encoder-Decoder SLM based on Public Datasets $\mathcal{P}$ and $\mathcal{R}$.}
  \FOR{each epoch $t \in [T]$}
       \STATE $\phi^{t+1} \gets \phi^{t} - \eta_\phi \nabla \mathcal{L}_{1} $.
    \ENDFOR

  \STATE \textcolor{gray}{$\triangleright$ Generate $p^p$ using the updated Encoder.}
  \STATE $p^p = \text{SLM}_{\text{Encoder}}(p)$.

  \STATE \textcolor{gray}{$\triangleright$ Generate perturbed rationales from LLM on the server.}
  \STATE $r^p = \text{LLM}(p^p)$.
  \STATE \textcolor{gray}{$\triangleright$ Decode perturbed rationales using the updated Encoder-Decoder SLM.}
  \STATE $r = \text{SLM}_{\text{Decoder}}(r^p)$.
 
  \STATE \textcolor{gray}{$\triangleright$ Multi-Task Training for Task-Specific SLM based on Datasets $\mathcal{D}$.}
  \FOR{each epoch $t \in [T]$}
       \STATE $\omega^{t+1} \gets \omega^{t} - \eta_\omega \nabla \mathcal{L}_{2} $. 
  \ENDFOR 
  
\end{algorithmic}
\end{algorithm}

\subsection{Privacy Preserving Prompt Encoder}
\label{sec: encoder}

Before the client transmits its raw prompts to the server-side LLM,  we need the privacy protection strategy to protect the raw prompts.
In this section, we propose two privacy protection strategies:
\begin{enumerate}

\item 

\textbf{Exponential Mechanism Encoder Strategy}.
In the first strategy, we utilize an exponential mechanism~\cite{mcsherry2007mechanism}, which satisfies the criteria for the $\epsilon$-DP. For detailed information about the exponential mechanism, please refer to Section \ref{sec:dp}.

Let us consider an Exponential Mechanism $M(\cdot)$.  Given a input prompt $p = \{x_i\}_{i=1}^S$ comprising $S$ tokens, a set $X$ encompassing all possible input tokens, and a set $Y$ of all potential output tokens,\textit{ the mechanism $M(\cdot)$ is applied to each input token} $x_i \in p$. If $x_i$ belongs to $X$, it is replaced with an output token $y_i$ from $Y$. Through this process, we obtain a perturbed prompt $p^p = \{y_i\}_{i=1}^S$.

\item 
\textbf{Adaptive Exponential Mechanism Encoder Strategy}.
\textit{The tokens within a prompt differ significantly in terms of their importance and degree of privacy.} Applying a uniform privacy budget  $\epsilon$ across all tokens may not lead to the most optimal solution. To further optimize the privacy-utility balance,  we propose an \textit{Adaptive Exponential Mechanism Encoder} strategy.
This strategy is built upon the first exponential mechanism. In the  \textit{Adaptive Exponential Mechanism Encoder} strategy, we utilize an Encoder-Decoder SLM specifically designed to encode raw prompts into perturbed prompts and subsequently decode perturbed rationales back into their original form. This strategy involves two training process: encoding training process and decoding training process. In this section, we mainly focus on encoding training process.

Initially, an encoding training process is required for the Encoder-Decoder SLM. Formally, let's denote a public dataset as $P=\left \{(p_i, p^{\epsilon}_i))  \right \}^N_{i=1} $, where $p_i$ represents raw private prompt, $p^{\epsilon}_i$ represents perturbed prompt generated using the first exponential mechanism with a privacy budget of $\epsilon$. In the encoding training process, we train the Encoder-Decoder SLM: $g_\phi(p_i) \to  p^{\epsilon}_i$. The details of  encoding training process is illustrated in Algorithm \ref{alg:fedcot}.

\textit{The Prompt Encoder objective can be formulated as follows:}
\begin{equation}\label{eq:encoder}
\begin{aligned}
     \mathcal{L}_{\text{Encoder}}(\phi;\mathcal{P}) = \mathbb{E}_{(p, p^\epsilon) \sim \mathcal{P}}\ell_{\text{CE}}(g_{\phi}(p), p^\epsilon)
\end{aligned}
\end{equation}
\textit{where $\ell_{\text{CE}}$ is the cross-entropy loss.}

\end{enumerate}

As illustrated in Figure\ref{fig:example},  we can observe an exemplary comparison between the original prompt and its perturbed prompt in Step 1 and Step 2.
This perturbed prompt serves as the new, privacy-enhanced input for further processing.

\subsection{Generating Perturbed Rationales from LLM}
When the server-side LLM receives the perturbed prompt, we leverage the Chain-of-Thought (CoT) prompting technique introduced by \cite{wei2022chain} to generate rationales from the LLM using this perturbed prompt. These generated rationales, which are also perturbed, are then transmitted to the client.
For instance, as illustrated in Figure \ref{fig:example}, given a perturbed prompt in the Step 2, the LLM generates perturbed rationales in the Step 3.

\subsection{Perturbed Rationales Decoder}
Once the client receives the perturbed rationales from the server-side LLM, it must initiate a "decoder" process  to decode the rationales. 
In this section, we also propose two strategies correspond to the two protection strategy of the prompt encoder module:

\begin{enumerate}
\item \textbf{Exponential Mechanism Decoder Strategy}. 
In the first decoding strategy, which corresponds to Exponential Mechanism Encoder strategy. Here, we utilize In-Context Learning (ICL) ~\cite{dong2024survey,tong2025inferdpt} with the Encoder-Decoder SLM to decode the perturbed rationales. 
we can input a sample $x_i=(p, p^p, r^p)_i$ into the Encoder-Decoder SLM to prompt the generation of rationales, where $p$ represents raw private prompt, $p^p$ represents perturbed prompt and $r^p$ represents perturbed rationales generated from LLM. $(p^p, r^p)_i$ can be viewed as an example for Encoder-Decoder SLM in ICL.
This allows the Encoder-Decoder SLM to generate rationales $r_i$ that are aligned with the original, unperturbed prompt. 

\item \textbf{Adaptive Exponential Mechanism Decoder Strategy}. 
In the second decoding strategy, which corresponds to Adaptive Exponential Mechanism Encoder strategy. The rationales decoder module also use the same the Encoder-Decoder SLM with Section \ref{sec: encoder}. 

Initially, a decoding training process is required for the Encoder-Decoder SLM.
Formally, let's denote a public dataset as $R=\left \{(x_i, r_i))  \right \}^N_{i=1} $, where $x_i$ represents an input, where $x_i=(p, p^p, r^p)_i$ , $p$ represents raw private prompt, $p^p$ represents perturbed prompt generated from Encoder-Decoder SLM,  $r^p$ represents perturbed rationales generated from LLM. $r_i$ represents the raw rationale of raw prompt $p$ generated from LLM. In the decoding training process, we train the Encoder-Decoder SLM: $g_\phi(x_i) \to  r_i$. The details of decoding training process is illustrated in Algorithm \ref{alg:fedcot}.

\textit{The Rationales Decoder objective can be formulated as follows:}
\begin{equation}\label{eq:decoder}
\begin{aligned}
     \mathcal{L}_{\text{Decoder}}(\phi;\mathcal{R})= \mathbb{E}_{(x,r) \sim \mathcal{R}}\ell_{\text{CE}}(g_{\phi}(x), r)
\end{aligned}
\end{equation}

Subsequently, once the decoding training process of Encoder-Decoder SLM is finished, we can input a sample $x_i=(p, p^p, r^p)_i$ into the SLM, where $r^p$ represents perturbed rationales generated from LLM. This allows the SLM to generate rationales $r_i$ that are aligned with the original, unperturbed prompt. 

We approach the training of the Encoder-Decoder SLM as a multi-task learning problem encompassing both the encoding and decoding training processes. 

\textit{The multi-task learning objective for the Encoder-Decoder SLM can be formulated as follows:}
\begin{equation}\label{eq:en-de}
\begin{aligned}
      \mathcal{L}_1 = \mathcal{L}_{\text{Encoder}} +  \mathcal{L}_{\text{Decoder}}
\end{aligned}
\end{equation}

\end{enumerate}

As illustrated in Figure\ref{fig:example},  we can observe an exemplary comparison between the perturbed rationales from LLM and its decoded rationales from SLM in Step 3 and Step 4.
It's worth noting that although the SLM has the ability to generate aligned rationales independently, the quality often falls short due to its limited capabilities. By leveraging the perturbed rationales, we effectively transfer the powerful capabilities of the server-side LLM to enhance the Encoder-Decoder SLM, thereby improving the overall quality of the generated rationales.

\subsection{Enhancing Task-Specific SLM via CoT Knowledge Distillation}

In our work, we undertake the training of the client's Task-Specific SLM tailored for text generation tasks. Initially, we elaborate on the prevalent framework for learning task-specific models. Leveraging this established framework, we enhance it by integrating rationales produced from the rationales decoder module into the training process. Formally, let's denote a dataset as $D=\left \{ (x_i, (y_i, r_i))  \right \}^N_{i=1} $, where $x_i$ represents an input, $y_i$ represents the associated expected output label, and $r_i$ is the corresponding desired rationale.

We conceptualize learning with rationales as a \textit{multi-task learning} problem. Specifically, we train the model  $f_\omega(x_i) \to  (y_i, r_i)$ to accomplish not just the prediction of task labels but also the generation of the corresponding rationales based on textual inputs.  
This multi-task training ensures that our model not only produces accurate predictions but also provides insightful justifications for its decisions. By doing so, we enhance the transparency and explainability of the model. 

\textit{The multi-task learning objective for the Task-Specific SLM can be formulated as follows:}
\begin{equation}\label{eq:mft}
\begin{aligned}
      \mathcal{L}_2 = \mathcal{L}_{\text{Label}} + \mathcal{L}_{\text{Rationale}}
\end{aligned}
\end{equation}
\textit{where $\mathcal{L}_{\text{Label}}$ is the label prediction loss:}
\begin{equation}\label{eq:ft}
\begin{aligned}
     \mathcal{L}_{\text{Label}}(\omega;\mathcal{D})= \mathbb{E}_{(x,y) \sim \mathcal{D}}\ell_{\text{CE}}(f_{\omega}(x), y)
\end{aligned}
\end{equation}
\textit{and $\mathcal{L}_{\text{Rationale}}$ is the rationale generation loss:}
\begin{equation}\label{eq:rationale}
\begin{aligned}
      \mathcal{L}_{\text{Rationale}}(\omega;\mathcal{D})= \mathbb{E}_{(x,r) \sim \mathcal{D}}\ell_{\text{CE}}(f_{\omega}(x), r)
\end{aligned}
\end{equation}
where $\ell_{\text{CE}}$ is the cross-entropy loss, $f_{\omega}(.)$ is the Task-Specific SLM model.

\subsection{Privacy Analysis of FedCoT}

The privacy-protection strategies in FedCoT implement a token-level Exponential Mechanism in feature space, adhering to the $\epsilon$-DP principles. This mechanism provides mathematically provable privacy guarantees at the token-level granularity, as extensively validated in privacy-preserving NLP research 
 \cite{yue2021differential, chen-etal-2023-customized, tong2025inferdpt}. Our experimental results further validate this approach: when privacy budget is low, the rationales generated from perturbed prompts show significantly lower similarity to those from original prompts, demonstrating the effectiveness of our privacy protection while acknowledging the inherent privacy-utility trade-off.

\section{Experiments}

\subsection{Setup}

We have established a scenario to evaluate the performance of the FedCoT framework across a range of text generation tasks. This setup involves a client-server architecture, where the client holds two downstream SLMs: an  \textit{Encoder-Decoder SLM}, which specializes in encoder-decoder functionalities and a \textit{Task-Specific SLM}, tailored for specific tasks. On the server-side, we host a LLM for more general and powerful text generation capabilities. 
Specifically, Table \ref{tab: modelsetting} outlines the detailed configurations of both the LLM and the SLMs. In our experimental setup, the \textit{Encoder-Decoder SLM} and \textit{Task-Specific SLM} are the identical architecture.

\begin{table}[ht]
    \centering
    \footnotesize
    \setlength{\tabcolsep}{2.5pt}
    \begin{tabular}{lccc}
        \toprule
        & & \multicolumn{2}{c}{\textbf{SLM}} \\
        \cmidrule(lr){3-4}
        \textbf{Setting} & \textbf{LLM} & \textbf{Encoder-Decoder}& \textbf{Task-Specific}  \\
        \midrule
        \multirow{1}{*}{Setting 1}& LLaMA3 70B& Pythia-1.4B& Pythia-1.4B\\
        \midrule
        \multirow{1}{*}{Setting 2}& Qwen1.5-14B& Qwen1.5-0.5B& Qwen1.5-0.5B\\
        
        \bottomrule
    \end{tabular}
\caption{ LLM and SLMs Setting of FedCoT.}
\label{tab: modelsetting}
\end{table}

\textbf{Datasets and Evaluation Metrics}. We conduct an evaluation of FedCoT on 4 QA datasets. Specifically, we include CommonsenseQA (CQA)~\cite{talmor2019commonsenseqa}, OpenBookQA (OBQA) ~\cite{mihaylov2018can}, BoolQ ~\cite{clark2019boolq}, ArcE~\cite{clark2018think}. For these datasets, we primarily use \textbf{Accuracy} as the evaluation metric. It's worth noting that in our experiments, all methods undergo zero-shot evaluation except FewShot(1-shot), and we use the \textit{lm-evaluation-harness} package~\cite{eval-harness}.

\textbf{Baselines}. 
Since we incorporate two distinct strategies in the prompt encoder and perturbed rationales decoder, we denote FedCoT method with the Exponential Mechanism Strategy as \textit{FedCoT-E} and FedCoT method with the Adaptive Exponential Mechanism Strategy as \textit{FedCoT-A}.  We conduct a comparative analysis to evaluate the performance of our FedCoT framework, which comprises both \textit{FedCoT-E} and \textit{FedCoT-A}.

These baselines included:
\begin{itemize}
    \item FewShot-LLM, which represents the few-shot capabilities of LLM  on the server; 
    \item FewShot-SLM, which represents the few-shot performance of SLM on the client; 
    \item Standalone, where the client fine-tunes its local model using its own private dataset;
    \item Non-Private, where the client send its raw local prompt to server, get rationales from LLM and fine-tunes its local model like FedCoT, but without privacy-preserving.
\end{itemize}

\subsection{Main Results}

In this section, we undertake a comparative analysis of the task performance of FedCoT. We assess both the FedCoT-E and FedCoT-A methods against other baselines on Task-Specific SLM under the privacy budget $\epsilon=3$. Our experiments encompass two model configurations: \textbf{\textit{Setting 1}} (LLM: LLaMA3-70B, Encoder-Decoder SLM \& Task-Specific SLM: Pythia-1.4B) and \textbf{\textit{Setting 2}} (LLM: Qwen1.5-14B, Encoder-Decoder SLM \& Task-Specific SLM: Qwen1.5-0.5B).

The results, as presented in Table \ref{tab:small-llm-performance}, clearly illustrate that both FedCoT-E and FedCoT-A exhibit significantly better performance when compared to FewShot-SLM and Standalone methods. Furthermore, FedCoT-A demonstrates notably superior performance compared to FedCoT-E. Specifically, take the model Setting 1 as an example, FedCoT-E surpasses the Standalone method by 4.3\%, 3.2\%, 7.1\%, and 5.1\% in the CQA, OBQA, BoolQ, and ArcE datasets, respectively. Meanwhile, FedCoT-A demonstrates even greater superiority, exceeding the Standalone method by 5.7\%, 4.6\%, 6.7\%, and 6\% across the same datasets.

\begin{table}[!ht]
\centering
\footnotesize
\setlength{\tabcolsep}{3.5pt}
\begin{tabular}{lccccc}
\toprule
 \textbf{Model} &\textbf{Method}  & \textbf{CQA} & \textbf{OBQA} & \textbf{BoolQ} & \textbf{ArcE}\\
\midrule
\multirow{7}{*}{
 \makecell[l]{\textbf{Setting 1}} 
 }
        & FewShot-LLM & 70.29 & 80.66 & 90.08 & 82.69 \\  
        \cmidrule(lr){2-6}  
        & FewShot-SLM & 21.19 & 26.60 & 52.11 & 28.91 \\  
         \cmidrule(lr){2-6}  
        & Standalone & 42.43 & 38.73 & 73.07 & 40.33 \\  
         \cmidrule(lr){2-6}  
       & Non-Private & 49.22 & 46.07 & 80.61 & 48.01 \\ 
        \cmidrule(lr){2-6}  
       & FedCoT-E & 46.70 & 41.93 & 80.02 & 45.42 \\  
     \cmidrule(lr){2-6}  
       & FedCoT-A & 48.10 & 43.30 & 79.77 & 46.34 \\  

    \midrule

    \multirow{7}{*}{

    \makecell[l]{\textbf{Setting 2}}
    }
    
     & FewShot-LLM & 80.9& 82.8& 85.2& 80.3\\  
        \cmidrule(lr){2-6}  
        & FewShot-SLM & 25.7& 28.6& 59.7& 40.7\\  
         \cmidrule(lr){2-6}  
        & Standalone & 55.7& 43.4& 78.4& 50.3\\  
         \cmidrule(lr){2-6}  
       & Non-Private & 59.3& 55.1& 80.5& 57.6\\ 
        \cmidrule(lr){2-6}  
       & FedCoT-E & 57.6& 50.8& 79& 52.6\\  
     \cmidrule(lr){2-6}  
       & FedCoT-A & 58.6& 53.1& 80.2& 56.5\\ 
         
\bottomrule

\end{tabular}
\caption{
We compare the performance of Task-Specific SLM trained with FedCoT-E ($\epsilon=3$)  and FedCoT-A ($\epsilon=3$) against the Task-Specific SLM trained using baseline methods. We consider two model settings: \textbf{Setting 1} (LLM: LLaMA3-70B, Encoder-Decoder SLM \& Task-Specific SLM: Pythia-1.4B) and \textbf{Setting 2} (LLM: Qwen1.5-14B, Encoder-Decoder SLM \& Task-Specific SLM: Qwen1.5-0.5B)} 

\label{tab:small-llm-performance}
\end{table}

\subsection{Performance Evaluation on various SLMs}
In this section, we extend the evaluation
of FedCoT’s effectiveness to encompass various client-side SLMs.
These SLMs include LLaMA2-1.3B~\cite{xia2024sheared}, Qwen1.5-1.8B~\cite{bai2023qwen}, and OPT-1.3B~\cite{zhang2022opt}. We have chosen LLaMA3-70B~\cite{dubey2024llama} as LLM.
Table \ref{tab:multi-slms} provides a clear illustration of how FedCoT(with $\epsilon$ = 3) consistently outperforms the Standalone method across various SLMs.

\begin{table}[!ht]
\footnotesize
\centering
\begin{tabular}{lcccc}
\toprule
\textbf{Dataset} & \textbf{Method} & \textbf{LLaMA2} &\textbf{Qwen1.5} & \textbf{OPT} \\
\midrule

\multirow{3}*{CQA} & Standalone& 61.5 & 57.8 & 56.42 \\  
\cline{2-5}  
~ & FedCoT-E& 63.03 & 60.30 & 57.55 \\  
\cline{2-5}  
~ & FedCoT-A& 64.27 & 62.21 & 60.18 \\  
\midrule
  
\multirow{3}*{OBQA} & Standalone& 47.53 & 52.60 & 40.93 \\  
\cline{2-5}  
~ & FedCoT-E& 51.73 & 56.40 & 49.13 \\  
\cline{2-5}  
~ & FedCoT-A& 49.8 & 57.20 & 48.4 \\  
\midrule
  
\multirow{3}*{BoolQ} & Standalone& 81.65 & 81.41 & 72.84 \\  
\cline{2-5}  
~ & FedCoT-E& 83.94 & 82.59 & 82.46 \\  
\cline{2-5}  
~ & FedCoT-A& 82.99 & 82.90 & 82.68 \\  
\midrule  
  
\multirow{3}*{ArcE} & Standalone& 40.33 & 55.58 & 45.92 \\  
\cline{2-5}  
~ & FedCoT-E& 54.11 & 61.07 & 49.67 \\  
\cline{2-5}  
~ & FedCoT-A& 54.66 & 62.43 & 50.69 \\  

\bottomrule
\end{tabular}
\caption{
We compare the performance of Task-Specific SLMs, which have been trained with FedCoT-E($\epsilon=3$)  and FedCoT-A($\epsilon=3$), against Standalone across various SLMs, including LLaMA2-1.3B, Qwen1.5-1.8B and OPT-1.3B.}
\label{tab:multi-slms}
\end{table}

\subsection{Ablation Study}

\textbf{Influence of Privacy Budgets}. 
We delve into the influence of privacy budgets on the performance of FedCoT. To ensure experimental consistency, we fix the model configuration to \textit{Setting 1} (as detailed in Table \ref{tab: modelsetting}) for all subsequent ablation experiments.
Table \ref{tab:compare_privacy} presents an overview of FedCoT's performance across a range of privacy budgets ($\epsilon = 1, 3, 5, 10$).

As the privacy budget $\epsilon$ increases, the performance of both FedCoT-E and FedCoT-A exhibits a notable uptick. Moreover, FedCoT-A consistently outperforms FedCoT-E under identical privacy budget conditions ($\epsilon$).
When compared alongside Table \ref{tab:small-llm-performance}, it becomes evident that with a privacy budget escalated to $\epsilon=10$, FedCoT-E surpasses the Standalone method by 5.6\%, 6.1\%, 6.3\%, and 6.8\% within the CQA, OBQA, BoolQ, and ArcE datasets, respectively. Similarly, FedCoT-A outperforms it by 4.3\%, 7.1\%, 6.8\%, and 7\%.
Notably, across all evaluated datasets, at a privacy budget of $\epsilon=10$, FedCoT attains performance levels comparable to Non-Private approaches, underscoring its proficiency and adaptability in striking a balance between privacy and utility.

\begin{table}[!ht]
\centering
\footnotesize
\begin{tabular}{lccccc} 
\toprule
\textbf{Method} & \textbf{$\epsilon$} & \textbf{CQA} & \textbf{OBQA} & \textbf{BoolQ} & \textbf{ArcE} \\
\midrule
\multirow{4}{*}{FedCoT-E}&1 & 45.63 & 42.13 & 78.91 & 44.84 \\
  \cmidrule(lr){2-6} 
  &3 & 46.70 & 41.93 & 80.02 & 45.42 \\
\cmidrule(lr){2-6} 
  &5 & 46.50 & 43.35 & 80.17 & 46.70 \\
 \cmidrule(lr){2-6} 
  &10 & 48.03 & 44.87 & 79.37 & 47.14 \\

\midrule
\multirow{4}{*}{FedCoT-A}&1 & 47.31 & 43.20 & 79.63 & 46.65 \\
  \cmidrule(lr){2-6} 
  &3 & 48.10 & 43.30 & 79.77 & 46.34 \\
 \cmidrule(lr){2-6} 
  &5 & 47.96 & 44.20 & 79.91 & 48.08 \\
 \cmidrule(lr){2-6} 
  &10 & 47.74 & 45.81 & 79.86 & 47.30 \\
 
\bottomrule
\end{tabular}
\caption{
Comparison of the performance of Task-Specific SLM trained with FedCoT-E and FedCoT-A across \textbf{different privacy budgets $\epsilon$}.
}
\label{tab:compare_privacy}
\end{table}

\textbf{Influence of Perturbed Rationales Decoding}.
We undertake an analysis to investigate the effects of perturbed rationales decoding on FedCoT when $\epsilon = 3$. Table \ref{tab:decoding_w_o} offers a comparison of FedCoT's performance, contrasting the results when perturbed rationales decoding is employed (FedCoT-E w/ and FedCoT-A w/) versus when it is not (FedCoT-E w/o and FedCoT-A w/o). 
Specifically, FedCoT-E w/ surpasses the FedCoT-E w/o by 2\%, 1.3\%, 1.5\%, and 0.6\% in the CQA, OBQA, BoolQ, and ArcE datasets, respectively. Meanwhile, FedCoT-A w/ demonstrates even greater superiority, exceeding the FedCoT-A w/o by 1.8\%, 1.6\%, 0.7\%, and 3\% across the same datasets. 
The findings unequivocally demonstrate that FedCoT exhibits superior performance when perturbed rationales decoding is utilized, as compared to when it is absent.

\begin{table}[ht]
    \centering
    \footnotesize
    \begin{tabular}{lccc}
        \toprule
        & & \multicolumn{2}{c}{\textbf{Decoding}} \\
        \cmidrule(lr){3-4}
        \textbf{Method} & \textbf{Dataset} & \textbf{w/}& \textbf{w/o}  \\
        \midrule
        \multirow{4}{*}{FedCoT-E}& CQA& 46.70& 44.79\\
        \cmidrule(lr){2-4}
        & OBQA& 41.93& 40.6\\
        \cmidrule(lr){2-4}
        & BoolQ& 80.02& 78.5\\
        \cmidrule(lr){2-4}
        & ArcE& 45.42& 44.78\\
        \midrule
        \multirow{4}{*}{FedCoT-A}& CQA& 48.10& 46.26\\
         \cmidrule(lr){2-4}
        & OBQA& 43.30& 41.7\\
        \cmidrule(lr){2-4}
        & BoolQ& 79.77& 79.06\\
        \cmidrule(lr){2-4}
        & ArcE& 48.08& 45.13\\
        \bottomrule
    \end{tabular}
    \caption{Comparison of Task-Specific SLM Performance in FedCoT: With vs. Without perturbed rationales decoding. 
    }  
    \label{tab:decoding_w_o}  
\end{table}

\textbf{Perturbed Rationales vs Original Rationales}.
We focus on analyzing the quality of the perturbed rationales ($r^p$) generated from the perturbed prompt of LLM based on FedCoT-E and FedCoT-A methods and compare them with the rationales ($r$) generated from raw prompt of the LLM. To evaluate the similarity between $r^p$ and $r$, we use \textit{TokenRatio} metric. A higher \textit{TokenRatio} indicates a greater degree of similarity between the perturbed and original rationales. 

\textbf{\textit{TokenRatio($r^{'}, r$)}}. 
This metric calculates the unique words($u$) in $r^{'}$ 
and counts how many of these words are also present in $r$, denoted as $i$. The \textit{TokenRatio} is then calculated as $i$ divided by the total number of unique words in $r^{'}$ ($|u|$).

As shown in Table \ref{tab:P-Rationales}, with an increase in the privacy budget $\epsilon$ and a corresponding decrease in perturbation, both the \textit{TokenRatio} of FedCoT-E and FedCoT-A have risen notably.  Furthermore, in most of tasks, the \textit{TokenRatio} of FedCoT-A is higher than that of FedCoT-E in the same level of  privacy budget $\epsilon$.  The experimental results confirm that the \textit{TokenRatio} observed in the perturbed rationales produced by both FedCoT-E and FedCoT-A, positively correlate with the privacy budget $\epsilon$. This suggests that as the privacy constraints are relaxed (higher $\epsilon$ values), the perturbed rationales become more similar to the original rationales.

\begin{table}[!ht]
\centering
\footnotesize
\begin{tabular}{lccccc} 
\toprule
\textbf{Method} & \textbf{$\epsilon$} & \textbf{CQA} & \textbf{OBQA} & \textbf{BoolQ} & \textbf{ArcE} \\
\midrule
\multirow{4}{*}{FedCoT-E}&1 &23.8 &33 &34.5 &26.7 \\ 
 \cmidrule(lr){2-6} 
 &3 &30.8 &45.26 &48.5 &44.7 \\ 
 \cmidrule(lr){2-6} 
 &5 &43.2 &66.3 &72.8 &67.4 \\ 
 \cmidrule(lr){2-6} 
 &10 &48.5 &75.8 &85.4 &74.5 \\ 
\midrule
\multirow{4}{*}{FedCoT-A}&1 &34.5 &37.9 &47.1 &20.7 \\ 
 \cmidrule(lr){2-6} 
 &3 &34.5 &49.5 &59.6 &30 \\ 
 \cmidrule(lr){2-6} 
 &5 &45.2 &69.6 &77.4 &36.2 \\ 
 \cmidrule(lr){2-6} 
 &10 &48.6 &76.12 &84.2 &38.6 \\ 
\bottomrule
\end{tabular}
\caption{We conduct a comparative analysis to assess the \textbf{perturbed rationales} produced by FedCoT-E and FedCoT-A methods against the \textbf{original rationales} that are directly generated from the raw prompt of the LLM. Metric used: TokenRatio.}
\label{tab:P-Rationales}
\end{table}

\textbf{Decoded Rationales vs Original Rationales}.
We delve into the quality analysis of the decoded rationales produced by the rationales decoder module based on FedCoT-E and FedCoT-A methods. We compare these decoded rationales against those generated directly from raw prompt of the LLM.
We utilize the \textit{TokenRatio} metric to assess their similarities.

As shown in Table \ref{tab:De-Rationales}, in contrast to FewShot-SLM, it becomes apparent that the decoded rationales' quality based on  FedCoT-E and FedCoT-A methods isn't solely reliant on the locally decoded SLM. The perturbed rationales crafted by the LLM indeed fulfill their intended purpose. When juxtaposed with Table \ref{tab:P-Rationales}, it's clear that at comparable $\epsilon$ levels, the \textit{TokenRatio} for the decoded rationales surpass those of the perturbed rationales in the FedCoT-E and FedCoT-A methods. This underscores the effectiveness of the rationales decoder module in the FedCoT-E and FedCoT-A methods.

\begin{table}[!ht]
\centering
\footnotesize
\begin{tabular}{lccccc} 
\toprule
\textbf{Method} & \textbf{$\epsilon$} & \textbf{CQA} & \textbf{OBQA} & \textbf{BoolQ} & \textbf{ArcE} \\
\midrule

 FewShot-SLM &- &42.9 &54.5 &35.8 &28.6 \\ \hline
 \multirow{4}{*}{FedCoT-E}&1 &36 &46.33 &44.13 &32.7 \\
 \cmidrule(lr){2-6} 
 &3 &39 &53.77 &53.1 &46 \\ 
 \cmidrule(lr){2-6} 
 &5 &44.8 &67.9 &73.9 &60.1 \\ 
 \cmidrule(lr){2-6} 
 &10 &48.4 &75.1 &85.4 &66.7 \\
 
\midrule

 \multirow{4}{*}{FedCoT-A}&1 &41.1 &60.36 &62.8 &42.19 \\ 
 \cmidrule(lr){2-6} 
 &3 &45.8 &65.35 &64.7 &42.99\\ 
 \cmidrule(lr){2-6} 
 &5 &50 &75.5 &72.9 &44.3 \\ 
 \cmidrule(lr){2-6} 
 &10 &53.3 &78.9 &76.6& 45.3 \\ 
 
\bottomrule

\end{tabular}
\caption{
We conduct a comparative analysis to assess the \textbf{decoded rationales} 
produced by FedCoT-E and FedCoT-A methods against the \textbf{original rationales} that are directly generated from the raw prompt of the LLM. Metric used: TokenRatio.}

\label{tab:De-Rationales}
\end{table}

\textbf{Outperforming Standalone with 50\% Data}.
We conduct an in-depth analysis to explore the influence of training data size on model performance. We compare the FedCoT method with the Standalone approach, varying the amount of training data used. Table \ref{tab:reduce} provides a clear illustration of how FedCoT(with $\epsilon=3$) achieves superior performance even with significantly fewer training samples compared to Standalone. More specifically, when trained on merely \textbf{50\%} of the complete CQA, OBQA, BoolQ, and ArcE datasets, both FedCoT-E and FedCoT-A either surpass or closely match the performance of Standalone method. 

\begin{table}[!ht]
\centering
\footnotesize
\begin{tabular}{lccccc}
\toprule
\textbf{Dataset} & \textbf{Method} & \textbf{25\%}   & \textbf{50\%} & \textbf{75\%} & \textbf{100\%} \\
\midrule

\multirow{3}*{CQA}   & FedCoT-E& 37.74 & 42.63 & 44.56 & 46.7  \\    
\cline{2-6}    
                     & FedCoT-A& 39.28 & 44.77 & 44.00 & 48.1  \\    
\cline{2-6}    
                     & Standalone & - & - & - & 42.43 \\    
\midrule   
\multirow{3}*{OBQA}  & FedCoT-E& 32.4 & 38.27 & 40.67 & 41.93 \\    
\cline{2-6}    
                     & FedCoT-A& 34.07 & 38.08 & 42.00 & 43.3  \\    
\cline{2-6}    
                     & Standalone & - & - & - & 38.73 \\    
\midrule   
\multirow{3}*{BoolQ} & FedCoT-E& 69.96 & 72.26 & 77.67 & 80.02 \\    
\cline{2-6}    
                     & FedCoT-A& 69.61 & 73.73 & 77.82 & 79.77 \\    
\cline{2-6}    
                     & Standalone & - & - & - & 73.07 \\    
\midrule   
\multirow{3}*{ArcE}  & FedCoT-E& 37.79 & 41.42 & 42.22 & 45.42 \\    
\cline{2-6}    
                     & FedCoT-A& 37.64 & 41.86 & 45.28 & 46.34 \\    
\cline{2-6}    
                     & Standalone & - & - & - & 40.33 \\

\bottomrule
\end{tabular}
\caption{
We compare the performance of Task-Specific SLM trained with FedCoT-E($\epsilon=3$)  and FedCoT-A($\epsilon=3$)  against Standalone, across a range of dataset sizes from 25\% to 100\%. The '-' indicates a method does not apply to the corresponding dataset sizes.}
\label{tab:reduce}
\end{table}

\section{Conclusions}

In this study, we introduce FedCoT, a federated framework designed to distill knowledge from LLMs to SLMs in resource-constrained environments. FedCoT facilitates secure knowledge transfer from LLMs to SLMs by leveraging perturbed prompts and rationales, thereby enhancing the performance of SLMs without compromising user privacy. We present two innovative privacy protection strategies, including an Adaptive Exponential Mechanism strategy, which effectively balance privacy preservation and the usability of rationales. Experiments on various text generation tasks demonstrate FedCoT’s ability to enhance SLM performance with LLM support while prioritizing data privacy.

\section*{Limitations}

While our proposed FeCoT framework demonstrates promising results in privacy-preserving knowledge transfer from LLMs to SLMs, it is important to acknowledge several considerations that could be addressed in future work. Firstly, the framework's performance benefits are contingent upon the server-side LLM's CoT reasoning capabilities. Although contemporary LLMs like GPT-4 and LLaMA exhibit strong reasoning skills, frameworks such as FedCoT may encounter limitations when deployed with less sophisticated LLMs. This suggests an opportunity for further research to enhance FedCoT's robustness against variability in LLM reasoning abilities.
Secondly, our evaluation primarily focused on LLaMA and Qwen as the server-side LLMs, with client-side SLMs including Pythia, LLaMA, Qwen, and OPT. While these models are representative of current state-of-the-art architectures, extending testing to a more diverse set of LLMs could provide deeper insights into FedCoT's generalizability.

\bibliography{ref}

\appendix

\section{FedCoT's Computational and Communication Overhead}
\label{sec:appendix-requirement}
FedCoT is designed to be efficient and scalable in resource-constrained environments. The communication overhead is minimal, with costs comparable to plaintext data transmission. Computational requirements are equivalent to standard SLM fine-tuning (SFT) on local tasks. Our experimental validation was conducted using NVIDIA V100 GPUs, demonstrating practical deployment feasibility.

\section{Rationales Generation through CoT}
\label{sec:appendix-COT}

We utilize the rationales data generated by server-side LLM through chain-of-thought (CoT)\cite{wei2022chain}\cite{hsieh2023distilling} technique to enhance the performance of the client's task-specific SLM. 
These rationales justify the predicted labels and serve as insightful guidance for training smaller and domain-specific models. Consider the following example: when asked “Question:A beaver is know for building prowess, their supplies come from where?
Answer Choices:
(a) british columbia
(b) body of water
(c) wooded area
(d) pay debts
(e) zoo”. Utilizing the chain-of-thought (CoT) technique, the LLM can generate intermediate rationales like, 
"The answer must be the place where beavers get their supplies. Of the above choices, only wooded areas have the supplies that beavers need.”  Such rationales bridge the gap between the input and the final answer, often encapsulating valuable task-related knowledge. This knowledge would traditionally require extensive data for smaller and task-specific models to acquire. Therefore, we harness these rationales as enriched training material for small language models, employing a multi-task training paradigm that encompasses both label prediction task and rationale prediction task.

\section{More on Experimental Details}
\label{sec:appendix-Experimental}

\subsection{Hyperparameter Settings}

\textbf{SLM Parameters.} During the training process for both the Encoder-Decoder SLM and the Task-Specific SLM, we specifically configured the parameters. We set the batch size to 32 and employed the AdamW optimizer. The maximum number of training steps ranged from 400 to 1500. Additionally, we assigned the values of 0.5 to both $\alpha$ and $\beta$.  Furthermore, the learning rates for $\eta_\phi$ and $\eta_\omega$ were established at 5e-5.

\subsection{Data Splitting}

For the datasets CQA/OBQA/BoolQ//ArcE/, all splits (training, validation, and test) were downloaded from HuggingFace \cite{lhoest-etal-2021-datasets}. During the training of the Encoder-Decoder SLM, we randomly divided the training data into two equal parts. One part was designated as the public dataset, while the other part was allocated as the private dataset for the client.

\subsection{Dataset Licenses}
For the datasets CQA/OBQA/BoolQ//ARC-E/ were downloaded from HuggingFace\cite{lhoest-etal-2021-datasets} and under Apache License, Version 2.0.

\subsection{Machine Configuration}

The experiments were conducted on machines equipped with 4 and 8 NVIDIA  V100 32G.

\end{document}